\documentclass[runningheads]{llncs}
\usepackage[T1]{fontenc}
\usepackage{graphicx,verbatim}
\usepackage{hyperref}
\usepackage{color}

\def\modelname{PathoPainter} 

\usepackage{booktabs}
\usepackage{amsmath, amsfonts, amssymb}

\usepackage{multirow}
\usepackage{pifont}
\usepackage{subcaption}
\usepackage{soul}
\def\onedot{.}
\def\eg{{\em e.g}\onedot} 
 
\def\vs{{\em vs}\onedot}

\newcommand{\subpara}[1]{\vspace{3pt} \noindent \textbf{#1.}}

\begin{document}
\title{\modelname: Augmenting Histopathology Segmentation via Tumor-aware Inpainting}

\author{Hong Liu\inst{*,1} \and
Haosen Yang\inst{*,2} \and
Evi M.C. Huijben\inst{1} \and
Mark Schuiveling\inst{3} \and
Ruisheng Su \inst{1} \and
Josien P.W. Pluim\inst{1} \and
Mitko Veta\inst{1}
}

\authorrunning{H.Liu et al.}

\institute{Eindhoven University of Technology, Eindhoven, The Netherlands 
\and
University of Surrey, London, UK \and
University Medical Centre Utrecht, Utrecht, The Netherlands \\
*Joint main authors \email{\{hong.liu.00408,haosen.yang.6\}@gmail.com}}

\maketitle            

\begin{abstract}
Tumor segmentation plays a critical role in histopathology, but it requires costly, fine-grained image-mask pairs annotated by pathologists.
Thus, synthesizing histopathology data to expand the dataset is highly desirable.
Previous works suffer from inaccuracies and limited diversity in image-mask pairs, both of which affect training segmentation, particularly in small-scale datasets and the inherently complex nature of histopathology images.
To address this challenge, we propose \textbf{\modelname{}}, which reformulates image-mask pair generation as a tumor inpainting task. 
Specifically, our approach preserves the background while inpainting the tumor region, ensuring precise alignment between the generated image and its corresponding mask.
To enhance dataset diversity while maintaining biological plausibility, we incorporate a sampling mechanism that conditions tumor inpainting on regional embeddings from a different image. 
Additionally, we introduce a filtering strategy to exclude uncertain synthetic regions, further improving the quality of the generated data.
Our comprehensive evaluation spans multiple datasets featuring diverse tumor types and various training data scales.
As a result, segmentation improved significantly with our synthetic data, surpassing existing segmentation data synthesis approaches, \eg, 75.69\% \(\rightarrow\) 77.69\% on CAMELYON16. The code is available at \url{https://github.com/HongLiuuuuu/PathoPainter}.


\end{abstract}
\section{Introduction}
\label{sec:intro}

Tumor segmentation is crucial in histopathology~\cite{wsisam,9008367,hooknet,QAISER20191,HAN2022102487}, but training effective segmentation models, especially in supervised scenarios, requires pixel-wise annotations on whole slide images (WSIs), which demands expert knowledge and significant effort due to their structural complexity and gigapixel scale. 
This often results in sparse and limited labeled datasets~\cite{diffinf}, as annotation of a single slide at high magnification can take an experienced pathologist hours.
This challenge, along with privacy regulations and ethical concerns~\cite{rieke2020federated,schwarz2019mri}, has led to the exploration of medical image-mask pair synthesis.

\begin{figure}[!t]
    \centering
    \includegraphics[width=\linewidth]{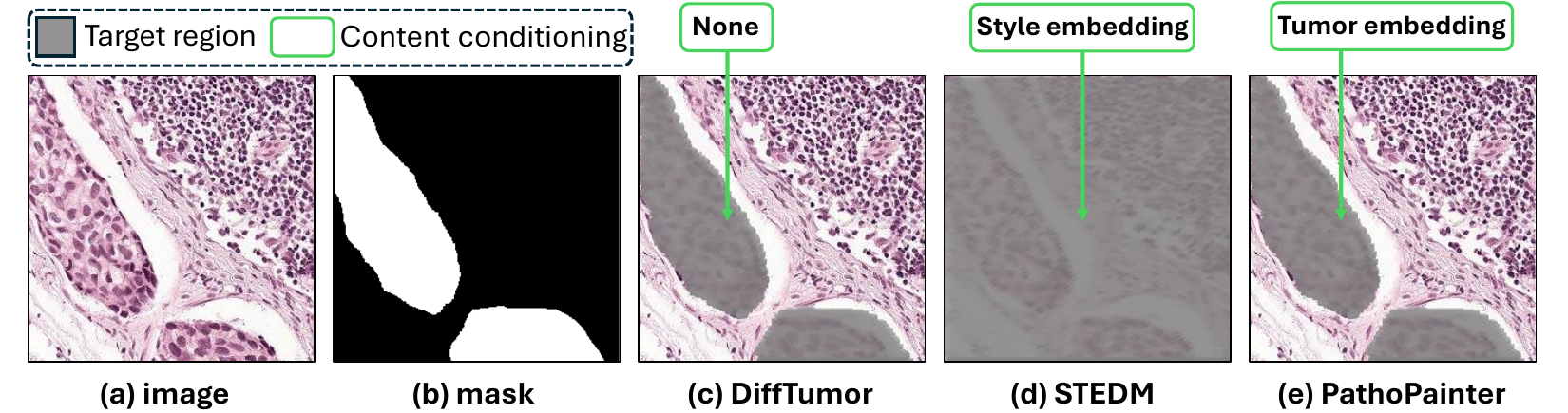}
    \caption{Illustration of various synthetic methods. The columns represent: (a) the original image \(x\), (b) the ground truth mask \(m\), (c) DiffTumor~\cite{qi}, which inpaints the tumor region using the mask as a condition without any content conditioning, (d) STEDM~\cite{style}, which generates the entire image with style conditioning, and (e) \modelname{} (ours), which inpaints the tumor region using the mask and tumor embeddings from a different image as conditions.}
    \label{fig:1}
\end{figure}

Recent advancements in generative models, particularly denoising diffusion probabilistic models (DDPMs)~\cite{ddpm}, have opened new avenues for conditional image synthesis~\cite{freemask,zhang2020learning}. 
DiffTumor~\cite{qi} (Figure~\ref{fig:1}(c)) augments computed tomography data via inpainting. However it tends to overfit in histopathology data due to the absence of semantic conditioning, failing to capture the intricate variations in appearance and structure.
Another approach~\cite{style} (Figure~\ref{fig:1}(d)) generates images conditioned on various styles to enhance  dataset diversity. However, this method often cause misalignment between the generated image and the ground truth mask, resulting in inaccuracies.
These issues negatively affect the training of segmentation models, which rely on high-quality image-mask pairs for optimal performance. Specifically in WSIs, the presence of subtle tissue patterns, intricate cellular details, and varying staining intensities make it crucial to maintain precise alignment between the image and its corresponding mask to ensure accurate segmentation~\cite{hooknet,wsisam,9008367,QAISER20191}.

In this paper, we propose \textbf{\modelname{}} (Figure~\ref{fig:1}(e)), which reformulates image-mask pair generation by inpainting images conditioned on regional tumor embeddings and masks.
Specifically, we apply a mask to retain non-tumor features and use regional tumor embeddings as content conditioning to generate the tumor region during training. 
During inference, we employ a sampling mechanism that selects tumor embeddings from a different image to generate the tumor region and maintain biological plausibility.
Synthetic image-mask pairs are used to train a segmentation model, where local inaccuracies, referred to as uncertain regions, can negatively impact learning.
We hypothesize that uncertain regions in synthetic images can be identified by their higher loss values when evaluated using a segmentation model pretrained on real data. To mitigate their impact, we propose a self-adaptive approach to filter out these regions from the synthetic images.
We summarize the contributions of this work as follows:

\begin{itemize} 

\item We propose \textbf{\modelname{}}, which reformulates image-mask pair generation by inpainting images conditioned on masks and regional tumor embeddings.
\item We introduce a conditioning mechanism that leverages regional tumor embeddings for content conditioning, combined with tumor embedding sampling and an uncertain region filtering strategy, to generate high-quality, accurate, and diverse image-mask pairs.
\item We thoroughly evaluate the effectiveness of \textbf{\modelname{}} in enhancing dataset diversity and boosting segmentation performance.
\end{itemize}

\section{Method}
\label{sec:method}

\begin{figure*}[t]
    \centering
    \includegraphics[width=0.98\linewidth]{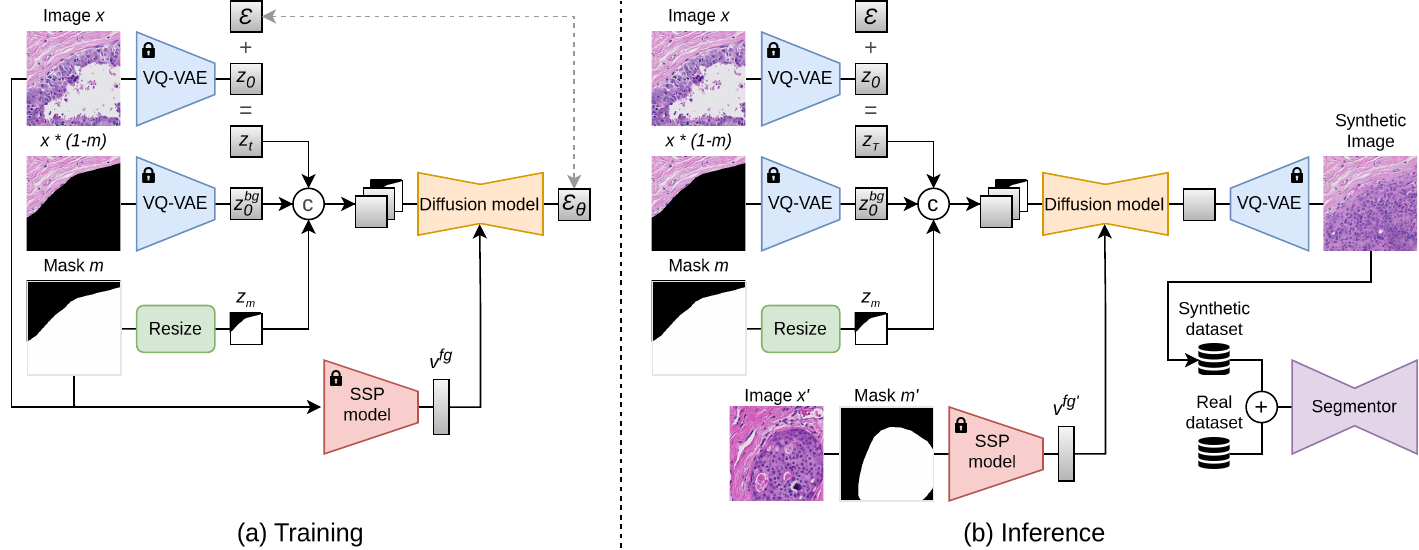}
    \caption{
    Overview of the \textbf{\modelname{}} framework.
    During (a) training, the input image \(x\) and the masked image \(x \odot (1 - m)\) are encoded using a pretrained VQ-VAE~\cite{vqvae} to obtain their latent representations \(z_0\) and \(z_0^{bg}\). These representations, concatenated with the resized mask \(z_m\), are fed into the diffusion model and conditioned on the embedding \(v^{fg}\) extracted by SSP model, which captures features exclusively from the foreground.
    To augment the training set for downstream segmentation tasks, we generate reliable image-mask pairs using the method in (b). We randomly select a foreground embedding \(v^{fg'}\) from the same tumor type as the target image. The U-Net denoiser then generates a new image-mask pair that contains tumors with characteristics distinct from those in the original image.}
    \label{fig:pipeline}
\end{figure*}

We propose a conditional diffusion model for histopathology image-mask pair synthesis (Figure~\ref{fig:pipeline}). First, we review DDPM (Section~\ref{sec:Preliminaries}), then introduce our Tumor-aware Conditional Stable Diffusion model: \textbf{\modelname} (Section~\ref{sec:tumor_painter}), and finally, describe the filtering mechanism to address uncertain regions (Section~\ref{sec:filter}).

\subsection{Preliminaries: Conditional Diffusion model}
\label{sec:Preliminaries}
The DDPM~\cite{ddpm} is a generative framework designed to learn a target data distribution. To overcome the computational challenges of training DDPMs on high-dimensional data, such as images, LDM~\cite{ldm} was introduced, enabling DDPM training in a latent space rather than the pixel space.
LDMs consist of three key components: 1) a Variational Autoencoder (VAE)~\cite{vae} that encodes an input image \(x\) into compact latent space \(z\), 2) a U-Net denoiser that learns to reverse the diffusion process by iteratively refining the noisy latent representations over \(T\) time steps, and 3) a conditioning mechanism to guide and control the generation process. The U-Net denoiser is trained to predict the Gaussian noise \(\epsilon\) added at each time step. Specifically, given the noisy latent \(z_t\) and a conditioning input \(c\), it learns to minimize the difference between the true noise \(\epsilon\) and the predicted noise \(\epsilon_\theta(z_t, c, t)\). The LDM training objective is defined as
\begin{equation}
\mathcal{L}_{ldm} =
    \mathbb{E}_{t\sim\mathcal{U}(0,T),\epsilon\sim \mathcal{N}(0,1)} 
    \left[
        \left\|
            \epsilon - \epsilon_\theta
            \left( 
                z_t, 
                c,
                t
            \right) 
        \right\|_2^2
    \right].
\end{equation}

\subsection{\modelname{}}
\label{sec:tumor_painter} 
\subpara{Training}
We first encode the input image \(x\) with a pretrained Vector quantized (VQ)-VAE~\cite{vqvae} to obtain its latent representation \(z_0\). Next, we encode the masked image \(x\odot(1-m)\) to derive a latent representation \(z_0^{bg}\) that reflects only background, or non-tumor, features. For compatibility, we resize the tumor mask \(m\) to match the dimensions of the latent representations and denote this resized mask as \(z_m\).

The input to the diffusion model is the concatenation of \(z_t\), \(z_0^{bg}\), and \(z_m\), where \(z_t\) is the initial latent representation \(z_0\) with Gaussian noise added \(t\) times. Specifically, by inputting the original image \(x\) and the tumor mask \(m\) to the self-supervised pretrained (SSP) model~\cite{hipt}, we extract a one-dimensional embedding \(v^{fg}\) with mask pooling technique~\cite{he2018maskrcnn}. 
This embedding represents regional tumor embedding, capturing information solely from the foreground (tumor) regions, as illustrated in the lower part of Figure~\ref{fig:pipeline}.
This process leads to a training objective that minimizes the difference between the added Gaussian noise \(\epsilon\) and the predicted noise \(\epsilon_\theta\) for a time step \(t\). The objective is defined as:
\begin{equation}
\mathcal{L}_{\modelname{}} =
    \mathbb{E}_{t\sim\mathcal{U}(0,T),\epsilon\sim \mathcal{N}(0,1)} 
    \left[
        \left\|
            \epsilon - \epsilon_\theta
            \left( 
                z_t,
                z_0^{bg},
                v^{fg},
                m,
                t
            \right) 
        \right\|_2^2
    \right].
\end{equation}
\subpara{Inference}
To augment the training set, we generate reliable image-mask pairs (Figure~\ref{fig:pipeline}(b)). Instead of using embeddings from the same image, our regional embedding sampling mechanism selects tumor embeddings \(v^{fg'}\) from a different image within the same tumor group. To ensure biological plausibility, we cluster all tumor embeddings from the training set into \(k\) groups using K-means~\cite{kmeans}. Given an image \(x\) for inpainting and another image \(x'\) providing \(v^{fg'}\) as content conditioning, this approach enhances the diversity of the dataset.

\subpara{Discussion}
Compared to directly generating an full image (Figure~\ref{fig:1}(d)), our approach results in more realistic and biologically plausible tumor regions that seamlessly blend into their surroundings. Additionally, by focusing on tumor regions within a given mask, our method reduces misalignment and unnatural artifacts common in full-image generation.

\subsection{Filtering uncertain synthetic regions}
\label{sec:filter}
We propose an adaptive filtering mechanism to exclude uncertain synthetic regions. 
Specifically, uncertain synthetic regions are less likely to be correctly identified by a well-trained semantic segmentation model.
Building on this observation, we filter synthetic pixels based on the following criterion: if a synthetic tumor pixel is not classified as a tumor pixel by a segmentation model pretrained on real dataset, it is marked as uncertain and excluded from the loss computation.
The uncertain regions are defined by the false negative pixels defined as \(FN = m - (m\cap P)\), where \(m\) is the ground-truth mask and \(P\) is the mask predicted by the pretrained segmentation model.

\section{Experiment}
\label{sec:experiment}

\subsection{Datasets}
\subpara{DCIS}
The first dataset is Ductal Carcinoma in Situ (DCIS)~\cite{dcis}, a non-invasive form of breast cancer that can progress to invasive ductal carcinoma. This dataset includes 116 WSIs with annotations from 109 patients. We utilize images at $10\times$ magnification and split the slides into 81 for training, 17 for validation, and 18 for testing. To evaluate our method's performance on smaller datasets, we randomly sample 8 and 20 WSIs.

\subpara{CATCH}
The second dataset is the Pan-tumor Canine Cutaneous Cancer Histology (CATCH) dataset~\cite{catch}. It consists of 350 WSIs with annotations from 282 patients, with 50 slides per tumor type. We use images at $5\times$ magnification and adhere to the official dataset split, with 245 slides for training, 35 for validation, and 70 for testing. To assess our method’s performance on reduced datasets, we randomly select 7 and 14 WSIs.

\subpara{CAMELYON16}
The final dataset is the CAMELYON16\setcounter{footnote}{0}\footnote{\url{https://camelyon17.grand-challenge.org/Data/}} challenge dataset for breast cancer. We use images at $10\times$ magnification and follow the official data split, with 111 annotated WSIs for training and 48 for testing. From the training set, we set aside 11 slides for validation. To evaluate our method on smaller subsets, we randomly select 10 and 20 WSIs.

\subsection{Implementation details}
Our implementation builds on publicly available latent diffusion code for histopathology, using masks to control tumor inpainting and incorporating regional tumor embeddings from a different image as content conditioning via an SSP model~\cite{hipt}.

\subpara{Latent Diffusion Model (LDM)}
In all experiments, we employ the LDM on the patches of $256\times256$ pixels. In particular, the pretrained VQ-VAE~\cite{ldm} with four downsampling blocks is used without any fine-tuning. For the U-Net denoiser, we use the model pretrained on ImageNet~\cite{Russakovsky2015}, as made available by~\cite{learnrepresent}, but fine-tuned it on our datasets for 20,000 training steps with a learning rate of $10^{-4}$. To fine-tune the U-Net, we used two NVIDIA A100 GPUs and trained with a batch size of 100 per GPU.

\subpara{Self-supervised pretrained (SSP) Model}
The SSP model employed in our work is HIPT~\cite{hipt}, which is fine-tuned on our training sets. Specifically, we fine-tune the $\text{VIT}_{256}$ model for 100 epochs using the AdamW~\cite{loshchilov2017decoupled} optimizer, with a batch size of 256 and a base learning rate of 0.0005. The first 10 epochs were used for warm-up to the base learning rate, followed by a decay phase governed by a cosine schedule. After fine-tuning, we extract features via mask pooling on the ViT feature map. 
We cluster all tumor embeddings from the training set into 10 groups using K-means~\cite{kmeans}.

\subpara{Segmentation model}
For the downstream experiments, we utilize the widely used segmentation model, the Swin-Unet, configured with a Tiny architecture~\cite{swinunet}. We perform binary segmentation to distinguish between foreground (tumor) and background (non-tumor). In experiments that involve training on synthetic images, we augment the real training set with synthetic images. The loss function consists of a weighted combination of cross-entropy loss and Dice loss, with both components assigned equal weights of 0.5.
We apply traditional data augmentation, including flipping, rotation, and color adjustments for training. The models are trained on an NVIDIA TITAN Xp GPU for 10 epochs, with batch sizes of 12 for Swin-Unet. The model with the lowest validation loss is selected for testing.
\subsection{Evaluation}
\begin{figure*}[t]
    \centering
    \includegraphics[width=\linewidth]{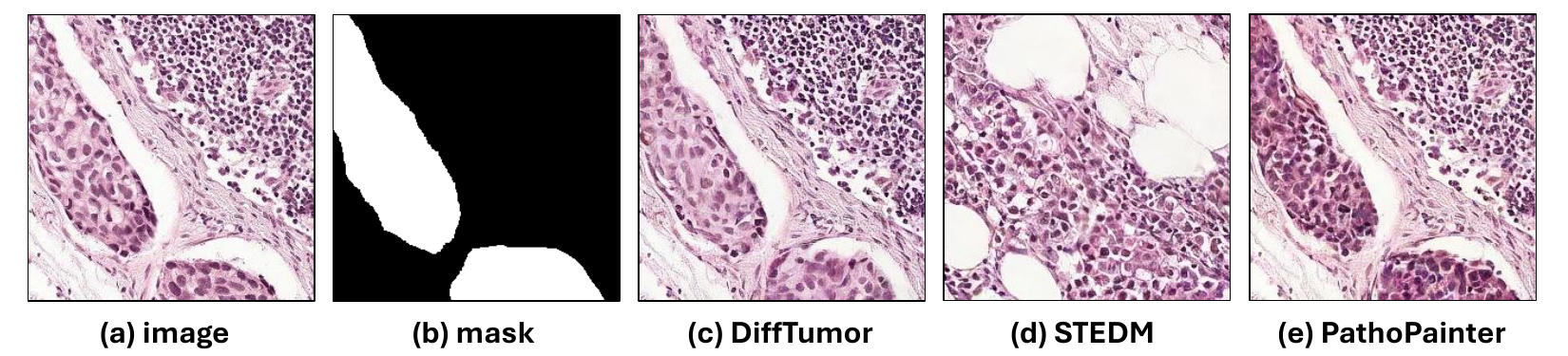}
    \caption{Visualization of generated images corresponding to the methods in Figure~\ref{fig:1}. The columns represent: (a) the original image \(x\), (b) the ground truth mask \(m\), and synthetic images generated by (c) DiffTumor~\cite{qi}, (d) STEDM~\cite{style}, and (e) \modelname{}.}
    \label{fig:vis_diff_datasets}
\end{figure*}
To validate our synthetic images, we first performed a qualitative assessment, examining image-mask alignment, deviations from the original images, and biological plausibility.
Following~\cite{freemask,qi}, we conducted a quantitative evaluation to verify the effectiveness of our method by assessing the performance of segmentation models trained on augmented datasets with synthetic images and tested on real data.
For this evaluation, we used the Intersection over Union (IoU) score for the foreground class and analyzed the variance of IoU scores across samples.

\subpara{Qualitative image generation results}
As shown in Figure~\ref{fig:vis_diff_datasets}, comparing our proposed \modelname{} with DiffTumor~\cite{qi}, the generated tumor region in (c) closely resembles the original image (a) in both staining and texture. However, in (e), the H\&E stain appears darker, and the tumor texture is denser while still maintaining biological plausibility.
When comparing \modelname{} with STEDM~\cite{style}, the full image generated in (d) no longer aligns with the ground truth mask (b), indicating inconsistencies. In contrast, \modelname{} preserves accurate alignment with the mask (b), ensuring better structural consistency.

Histopathological images generated using \modelname{} effectively preserve the mask, ensuring the generation of precise, high-quality image-mask pairs. When integrated with real images, these synthetic samples enhance the augmented dataset, providing robust and well-aligned training data that significantly improve segmentation model performance. 
leveraging tumor embedding of a different image, \modelname{} introduces greater diversity into the dataset while maintaining biological plausibility, further enriching the training process.
\begin{table*}[t]
    \caption{Segmentation results across different datasets and slide numbers.
    }
    \centering
    \renewcommand{\arraystretch}{1.0}  % Reduce row spacing
    \setlength{\tabcolsep}{3pt}  % Reduce column spacing
    \begin{tabular}{llcccc}
        \toprule
        % \textbf{Dataset} & \textbf{Synthetic Data} & \multicolumn{2}{c}{\textbf{8 WSI}} & \multicolumn{2}{c}{\textbf{20 WSI}} \\
        \textbf{Dataset} & \textbf{Synthetic Data} & \textbf{IoU (\%)} & \textbf{Variance} &
        \textbf{IoU (\%)} & \textbf{Variance} \\
        \midrule
        % \cmidrule(lr){3-4} \cmidrule(lr){5-6}
        % & & \textbf{IoU (\%)} & \textbf{Variance} & \textbf{IoU (\%)} & \textbf{Variance} \\
        & & \multicolumn{2}{c}{\textbf{8 WSI}} & \multicolumn{2}{c}{\textbf{20 WSI}} \\
        \cmidrule(lr){3-4} \cmidrule(lr){5-6}
        \multirow{4}{*}{\textbf{DCIS}}  
        & None & 62.33 & 0.1543 & 64.85 & 0.1432 \\
        & DiffTumor~\cite{qi} & 62.43 & 0.1529 & 65.03  & 0.1430\\
        & STEDM~\cite{style} & 61.98 & 0.1541  & 64.88 & 0.1492 \\
        & PathoPainter & \textbf{63.67} & \textbf{0.1510} & \textbf{66.09} & \textbf{0.1412} \\
        \midrule
        \multirow{4}{*}{\textbf{CATCH}}  
        & & \multicolumn{2}{c}{\textbf{7 WSI}} & \multicolumn{2}{c}{\textbf{14 WSI}} \\
        \cmidrule(lr){3-4} \cmidrule(lr){5-6}
        & None & 81.88 & 0.1017 & 84.77 & 0.0888 \\
        & DiffTumor~\cite{qi} & 82.01 & 0.0984 & 84.92 & 0.0849 \\
        & STEDM~\cite{style} & 82.23  & 0.1123 & 84.87 & 0.0852\\
        & PathoPainter & \textbf{83.10} & \textbf{0.0941} & \textbf{85.82} & \textbf{0.0809} \\
        \midrule
        \multirow{4}{*}{\textbf{CAMELYON16}}  
        & & \multicolumn{2}{c}{\textbf{10 WSI}} & \multicolumn{2}{c}{\textbf{20 WSI}} \\
        \cmidrule(lr){3-4} \cmidrule(lr){5-6}
        & None & 70.96 & 0.1516 & 75.69 & 0.1348 \\
        & DiffTumor~\cite{qi} & 71.20 &0.1542&76.21 &0.1321 \\
        & STEDM~\cite{style} & 71.08 & 0.1551 & 75.92 & 0.1338 \\
        & PathoPainter & \textbf{72.42} & \textbf{0.1501} & \textbf{77.69}  & \textbf{0.1243} \\
        \bottomrule
    \end{tabular}
    \label{tab:segmentation_results_comp}
\end{table*}

\subpara{Segmentation results}
We select DiffTumor~\cite{qi} and STEDM~\cite{style} as representative state-of-the-art methods for segmentation data synthesis.
Table~\ref{tab:segmentation_results_comp} presents the performance of the Swin-Unet model across various datasets and training data scales when incorporating synthetic training data from different models.
Across all datasets (DCIS, CATCH, and CAMELYON16), incorporating our synthetic data consistently improves IoU scores compared to using real data alone (\eg, 63.67\% \vs~62.33\% with 8 slides on DCIS).  
In contrast, STEDM results in a performance drop due to inaccurate layout generation. Similarly, TumorDiff shows only marginal improvement, as its tumor conditioning mechanism limits diversity.  
Furthermore, our method consistently outperforms other synthesis approaches, achieving the best overall performance across all datasets.

\begin{table*}[t]
\centering
\caption{Analysis of the effects of different design choices on segmentation performance on DCIS~\cite{dcis} with the Swin-Unet.  From left to right are the ablations on:
(a) Comparison of embeddings from the SSP model using the full image vs. regional tumor.
(b) Impact of embedding source: current image vs. other images.
(c) Effectiveness of the filtering strategy for synthetic data.}
\renewcommand{\arraystretch}{1.2} % Improve row spacing
\setlength{\tabcolsep}{6pt} % Adjust column spacing

\resizebox{\textwidth}{!}{ % Force all tables to fit within one row
\begin{tabular}{c c c}
% First subtable
\begin{tabular}{l|c}
\hline
\textbf{SSP Embedding} & \textbf{IoU (\%)}  \\
\hline
Full image & 61.28 \\
Regional tumor & \textbf{63.67} \\ 
\hline
\end{tabular} 
& % This ensures tables are side by side
% Second subtable
\begin{tabular}{l|c}
\hline
\textbf{Embedding Source} & \textbf{IoU (\%)}  \\
\hline
Self & 61.87 \\ 
Other & \textbf{63.67} \\ 
\hline
\end{tabular} 
& % This ensures tables are side by side
% Third subtable
\begin{tabular}{l|c}
\hline
\textbf{Filter} & \textbf{IoU (\%)} \\
\hline
w/o. & 61.62  \\  
w/. & \textbf{63.67} \\ 
\hline
\end{tabular}  
\end{tabular}
} % End resizebox

\label{tab:ab}
\end{table*}

\subsection{Ablation Studies}
We conduct ablation studies on 8 slides of DCIS dataset.

\paragraph{Effectiveness of regional tumor embedding.}
Table~\ref{tab:ab}(a) evaluates the impact of regional tumor embeddings, extracted using the SSP model and mask, on segmentation performance. Conditioning on regional tumor embeddings improves the IoU from 61.28\% to 63.67\% compared to full-image embeddings, demonstrating its effectiveness in capturing more precise semantic features directly relevant to tumor regions.

\paragraph{Effectiveness of embedding source.}
Table~\ref{tab:ab}(b) shows the impact of embedding sources on model performance. Utilizing embeddings from a different image demonstrates clear advantages by introducing diversity into the training data.

\paragraph{Effectiveness of filter strategy.}
Table~\ref{tab:ab}(c) demonstrates the effect of the filtering strategy, explained in Section~\ref{sec:filter}, on segmentation performance. Applying the proposed filtering strategy results in a substantial improvement in IoU from 61.62\% to 63.67\%. This demonstrates that adaptively filtering out uncertain synthetic regions effectively enhances the quality of the training data, thereby improving segmentation accuracy and ensuring more consistent results.

\section{Conclusion}
\label{sec:conclusion}
In this work, we proposed \modelname{}, a novel framework for generating synthetic histopathology image-mask pairs to address the challenges of limited annotated datasets in tumor segmentation. By preserving the background and conditioning image generation on tumor region, our method ensures alignment between the image and its corresponding mask. Additionally, we introduced a tumor embedding sampling mechanism to enlarge dataset diversity while preserving biological plausibility. To further improve data quality, we developed a simple yet effective filtering strategy to suppress uncertain synthesized regions. Comprehensive evaluations across multiple datasets demonstrated the effectiveness of our approach in enriching dataset diversity and enhancing segmentation performance, highlighting its potential to advance pathology image analysis.

\bibliographystyle{splncs04}
\bibliography{mybib}

\end{document}